# The Potential of Wearable Sensors for Assessing Patient Acuity in Intensive Care Unit (ICU)

Jessica *Sena, Mohammad Tahsin Mostafiz*[1]*, Jiaqing Zhang*[2]*, Andrea Davidson, Sabyasachi Bandyopadhyay, Ren Yuanfang, Tezcan Ozrazgat-Baslanti, Benjamin Shickel, Tyler Loftus, William Robson Schwartz, Azra Bihorac, and Parisa Rashidi*

*Abstract*—Acuity assessments are vital in critical care settings to provide timely interventions and fair resource allocation. Traditional acuity scores rely on manual assessments and documentation of physiological states, which can be time-consuming, intermittent, and difficult to use for healthcare providers. Furthermore, such scores do not incorporate granular information such as patients' mobility level, which can indicate recovery or deterioration in the ICU. We hypothesized that existing acuity scores could be potentially improved by employing Artificial Intelligence (AI) techniques in conjunction with Electronic Health Records (EHR) and wearable sensor data. In this study, we evaluated the impact of integrating mobility data collected from wrist-worn accelerometers with clinical data obtained from EHR for developing an AI-driven acuity assessment score. Accelerometry data were collected from 86 patients wearing accelerometers on their wrists in an academic hospital setting. The data was analyzed using five deep neural network models: VGG, ResNet, MobileNet, SqueezeNet, and a custom Transformer network. These models outperformed a rule-based clinical score (SOFA= Sequential Organ Failure Assessment) used as a baseline, particularly regarding the precision, sensitivity, and F1 score. The results showed that while a model relying solely on accelerometer data achieved limited performance (AUC 0.50, Precision 0.61, and F1-score 0.68), including demographic information with the accelerometer data led to a notable enhancement in performance (AUC 0.69, Precision 0.75, and F1-score 0.67). This work shows that the combination of mobility and patient information can successfully differentiate between stable and unstable states in critically ill patients.

## I. INTRODUCTION

Acuity refers to the severity of a patient's condition, which is concomitant with the priority assigned to patient care in a critical care setting. Patients in the intensive care unit (ICU) exhibit volatile physiological patterns and the potential for developing life-threatening conditions in a short span of time. Therefore, the timely recognition of evolving illness severity is of immense value in the ICU. Swift and precise assessments of illness severity can identify patients requiring the administration of immediate life-saving interventions [9]. Furthermore, these assessments can guide collaborative decision-making involving patients, healthcare providers, and families in determining care goals and optimizing resource allocation [1]. Patient acuity is a foundational concept in critical care that ensures patient needs are met with precision, safety, and efficiency. Accurate acuity assessments are crucial for guiding clinical interventions, optimizing staffing ratios, and ensuring the presence of adequately trained personnel to address the needs of high-acuity patients [7, 33]. From management and fiscal perspectives, an accurate understanding of in-patient acuity levels permits effective budgeting and resource allocation [30].

Traditional, manual, threshold-based scoring systems such as the Acute Physiology and Chronic Health Evaluation (APACHE) [2], the Simplified Acute Physiology Score (SAPS) [3], Sequential Organ Failure Assessment (SOFA) [4], Modified Early Warning Score (MEWS) [5] and others, have been developed to predict the risk of mortality in ICU patients and, by extension, gauge the complexity of their care needs [2]. These tools evaluate physiological parameters, laboratory results, and other pertinent clinical information.

Recent studies in clinical informatics have validated the effectiveness of automated machine learning methods by utilizing comprehensive data from Electronic Health Record (EHR) systems. Advanced algorithms using deep learning techniques have proven superior to conventional bedside severity evaluations in predicting in-hospital deaths, an indirect measure of immediate patient acuity. However, these systems are limited to physiological data captured within the EHR and neglect other significant

---
[1, 2] Mohammad Tahsin Mostafiz and Jiaqing Zhang contributed equally to this work.

aspects impacting the patient, such as mobility and functional status [9].

To overcome these limitations, Davoudi et al. [8] explored the benefits of augmenting traditional ICU EHR-based data with continuous and pervasive sensing technology. Their study combined EHR data with patient-worn accelerometer sensors, room light, and sound sensors, and a patient-facing camera to classify delirium in a small cohort of ICU patients. Inspired by the positive impact of these novel clinical data streams, Shickel et al. [9] proposed to augment EHR data with continuous activity measurements via wrist-worn accelerometer sensors to predict hospital discharge as a proxy for acuity. Their results showed that these augmented models were better at capturing illness severity.

In this work, we propose to evaluate the viability of using accelerometer and EHR data to assess patients' acuity directly. Following the same acuity phenotyping approach proposed by Ren et al. [6], the goal is to discern the patient's state as stable or unstable. To achieve this, we have developed an end-to-end deep learning pipeline based on sensor and EHR data (Figure 1).

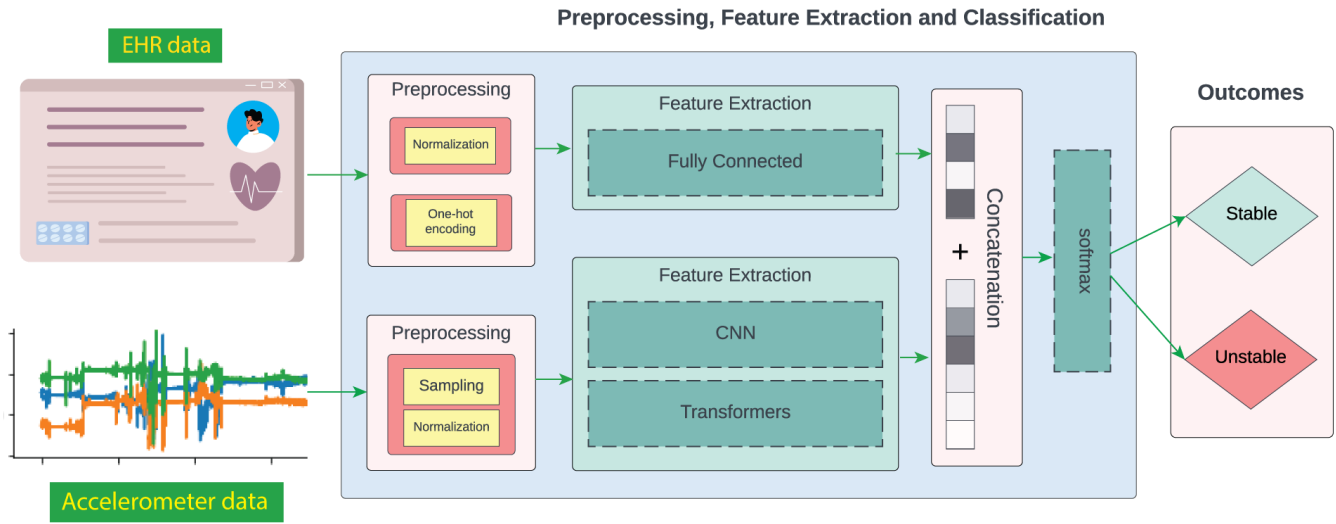

**Fig 1**. The proposed approach is an end-to-end neural network system that leverages accelerometer and EHR data to assess patient acuity, discerning between stable and unstable states.

In our approach, we evaluated five different neural network architectures, namely, VGG, ResNet, MobileNet, SqueezeNet, and a custom Transformers network, as both Convolutional Neural Networks (CNNs) and Transformers architectures are well-researched in the sensor-based human activity recognition field [18-23]. Due to their architectural advantages, CNNs are particularly adept at extracting features from accelerometer data. Their design promotes local connectivity, making them proficient in recognizing short sequences of time-series data and specific motion patterns. The shared weights in CNNs enable them to detect patterns regardless of their position in the sequence, while the hierarchical structure allows for extracting both simple and complex movement patterns. On the other hand, Transformers are advantageous for processing accelerometer data due to their self-attention mechanism, which aptly captures dependencies in time-series data. It allows the model to weigh the importance of different elements in a sequence when generating outputs [17]. Consequently, each patient's movement can be contextualized in relation to the other movements within a time window, directing the network's attention to the key movement patterns for assessing the patient's condition.

## II. MATERIALS AND METHODS

### A. Study Cohort

The data used in this research were sourced from adult patients admitted to the surgical ICUs at the University of Florida (UF) Shands Hospital main campus in Gainesville in compliance with all relevant federal, state, and university laws and regulations. Approval for the study was granted by the University of Florida Institutional Review Board under the numbers IRB201900354 and IRB202101013. Before enrolling patients in the study, written informed consent was obtained from all participants. In cases where patients could not provide informed consent, consent was obtained from a legally authorized representative (LAR) acting on their behalf. Eligible participants were individuals aged 18 and above admitted to the ICU and expected to remain there for at least 24 hours. Patients who could not provide LAR or self-consent were expected to be transferred or discharged from the ICU within 24 hours, and those necessitating contact or isolation precautions were excluded. Also excluded from this study were patients who expired within 24 hours of recruitment or from whom we could not collect accelerometer data. It includes the presence of intravenous lines, wounds, other hospital equipment, or the patient's choice to opt out of accelerometer placement.

Datasets were acquired from 86 critically ill patients between June 2021 and February 2023. Figure 1 depicts the data sources: Electronic Health Records (EHR) and accelerometer readings. Patients wore Shimmer3 [31] or ActiGraph wGT3X-BT [32] accelerometers on their wrists. Accelerometer readings were taken for a maximum of 7 days or until the patient's discharge from the ICU, whichever came first. Data from ActiGraph devices were retrieved using the ActiLife toolbox[3]. In contrast, data from the Shimmer device was uploaded and exported to a secure server via the Consensys software[4].

UF's Integrated Data Repository service extracted clinical data relevant to the patient's acuity state from the EHR. This information included age, sex, race, height, weight, and length of stay referred in this work as *demographics*, and physiological signals like blood pressure, heart rate, oxygen saturation (SpO2), pain score, Braden score [28], and cognitive status (whether the patient was in a coma, experiencing delirium, or had normal cognitive status) referred as *clinical data*.

### B. Data Processing

Acuity was assessed using phenotyping every four hours, and to capture relevant data leading up to each assessment, we established as a sample a 4-hour window that concluded immediately before the acuity evaluation. To phenotype the patient state as stable or unstable, we applied the method devised by Ren et al. [6], which determines transitions in acuity status within the ICU. For every 4 hours leading up to the assessment, patients—excluding those who had passed away or were already discharged alive—were identified as unstable or stable. A patient was labeled as unstable if they required any of the following life-supportive therapies: vasopressors, mechanical ventilation, continuous renal replacement therapy, or a massive blood transfusion (defined as at least ten units in the previous 24 hours). If none of these conditions were met, the patient was considered stable.

To address the varying sampling frequencies of the accelerometer data, we normalized all data to a consistent 10 Hz sampling frequency, as suggested by Antonsson and Mann [24]. They demonstrated that human activities typically range from 0-20 Hz, with 98% of the Fast Fourier Transform (FFT) power in accelerometer data falling within the 0-10 Hz range. This normalization ensures uniformity in the input data rate, facilitating more accurate analysis. Additionally, accelerometer values were rescaled to a range of [0, 1] to accommodate the requirements of the deep learning methods evaluated in our study. Similarly, numerical demographic data, such as age, was rescaled to the [0, 1] range, while categorical demographic information like sex and race was one-hot encoded. All clinical data consisted of time series captured within 4-hour windows, each varying in length. For our analysis, we transformed these time series into aggregated values. Specifically, numerical clinical data were computed as averages, while categorical clinical data were consolidated using majority voting in the 4-hour window.

### C. Deep Learning Models

The methods employed in this work were VGG [11], ResNet [12], MobileNet [13], SqueezeNet (SENet) [14], and a custom Transformer-based network [15]. The selection was grounded in their capabilities: VGG and ResNet for their depth that equips them to facilitate precise decision-making for the patients in the ICU, MobileNet, and SENet for their ability to maintain efficiency with a small number of parameters compared to ResNet and VGG, making them a suitable choice for deployment on constrained devices and reducing the decision-making latency which is crucial in the ICU setting. Transformer was selected for their attention features that empower the model to identify critical factors and events, aiding in precise decision support in the ICU setting. VGG, ResNet, MobileNet, and SENet were initially designed for image classification; thus, an adaptation of their architecture was necessary to suit accelerometer data. We tailored the original models to process 1D times-series while preserving the fundamental layer-wise structure and defining characteristics. It entailed replacing 2D Convolution, Average Pooling, and Max Pooling layers with their 1D counterparts and adjusting input channel configurations to match our data dimensions. For ResNet, SqueezeNet, and MobileNet, we retained essential components such as residual blocks (in ResNet), Squeeze-and-Excitation blocks (in SqueezeNet), and Depthwise Separable Convolution blocks (in MobileNet), with modifications primarily consisting of substituting 2D convolution and pooling filters with their 1D counterparts and updating channel parameters. The fully connected layers were kept unchanged. To further aggregate clinical and demographic features to the classification pipeline, we concatenated them with the dense features extracted from the fully connected layer.

In contrast, Transformer architectures are innately suited for sequence data processing. In our methodology, we extracted sequential feature embeddings from raw accelerometer sensor data using a feature embedding convolution layer. We then fed the extracted features, followed by adding positional encodings to capture the temporal order of the data into a Transformer encoding layer to learn the context and dependencies among the features. We further processed the extracted contextual features through another set of convolution and fully connected layers to enable our downstream classification tasks. We concatenated clinical and demographic features with the dense features extracted from the fully connected layer, similar to the approach adopted in earlier models. The model architecture is demonstrated in Figure 2.

---

[3] https://actilife.theactigraph.com/actilife
[4] https://www.consensys.net

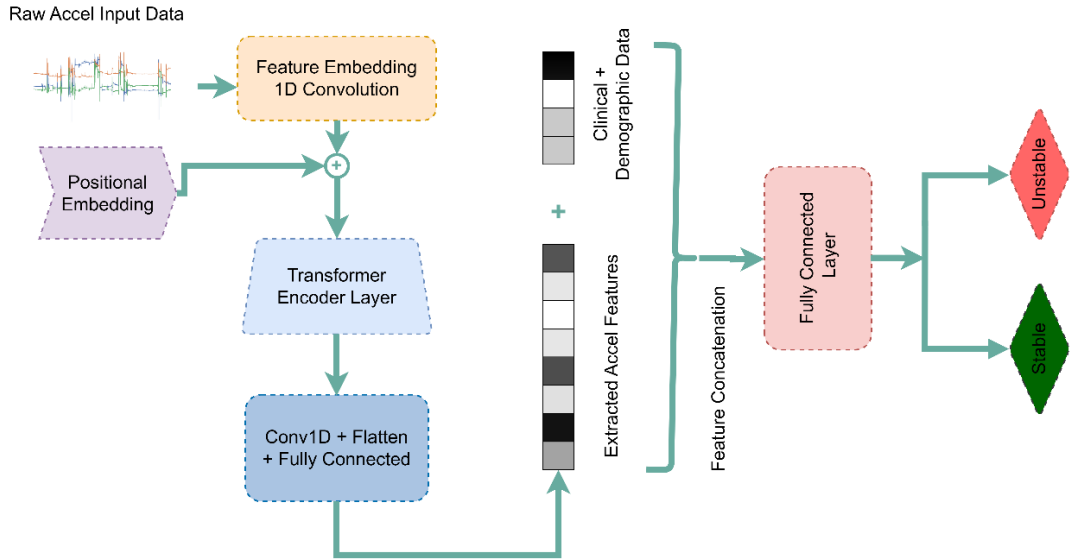

**Fig 2.** The proposed Transformer architecture for the acuity state classification task. Here, Accel stands for Accelerometer Sensor, Conv1D stands for 1D Convolution, Flatten stands for 2D to 1D flattening Layer, and Fully Connected stands for fully connected layer.

### *D. Experiments*

In assessing the performance of the deep learning models, we implemented an evaluation protocol to ensure robust and transparent assessment. The primary components of this protocol included a k-fold cross-validation approach in conjunction with the holdout approach, generating three distinct datasets: training, validation, and test datasets. The holdout approach was initially used to divide the dataset into development and test sets in a 70%/30% split. The k-fold cross-validation approach was then applied to the development dataset to generate the training and validation sets further to protect against potential overfitting. It entailed random partitioning of the development set into k equally sized subsets. A single subset was then retained as the validation set, while the remaining k-1 subsets were utilized for training. The process was repeated k times, using a different subset as the validation data, resulting in k different models and performance estimates. This method ensured that every development data point was used in a validation set once and in a training set k-1 times, giving a more robust performance estimate, especially in datasets of limited size, as is our case. The models were trained and validated on the development dataset to find the optimal hyperparameters. Once the models were finalized, the final models were evaluated on the test set, offering an estimation of their performance on unseen data.

We partitioned the dataset into three folds and a separate holdout test set based on individual patients. Each patient contributed multiple data points, which we termed as samples. Figure 3 shows the patient and sample distribution for our dataset.

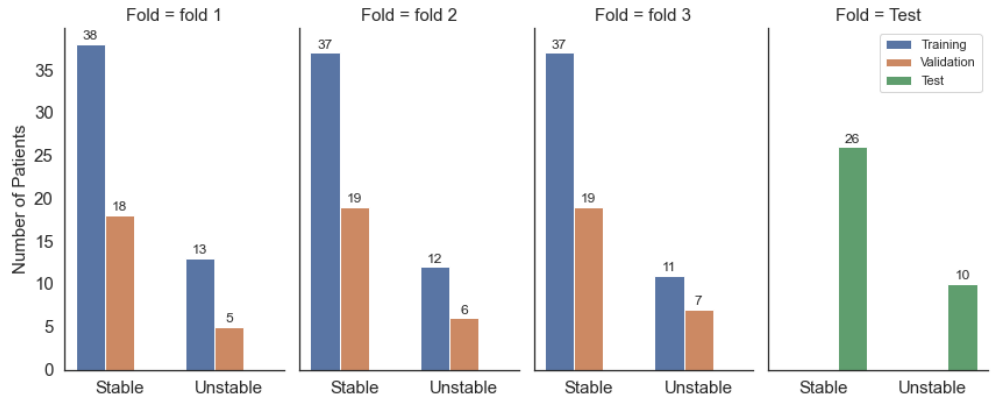

(a) Patient distribution

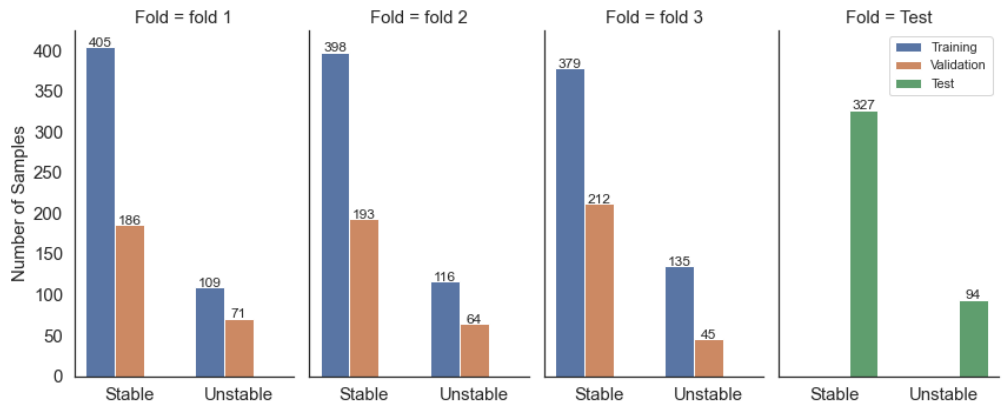

(b) Sample distribution

**Fig 3**. Distribution of patients and samples distribution.

Additionally, we incorporated the Sequential Organ Failure Assessment (SOFA) score as a rule-based scoring system into our evaluation process. The SOFA score, well-established in assessing patients in intensive care units (ICUs), provides an objective and standardized means of tracking a patient's condition over time. These properties make the SOFA score an indicator for the acuity state assessment task. To measure the acuity states, we normalized the SOFA scores within the range of [0, 1]. We treated these normalized scores as probability values and utilized the Youden index [29] to determine the optimal threshold for classifying the normalized scores and generating predictions. These predictions served as a baseline for comparison against the deep learning algorithms.

During the k-fold cross-validation process, we utilized Optuna [25] to search over the hyperparameters rather than a traditional grid search. Optuna reduces the runtime by pruning fewer promising trials during runtime. For every set of hyperparameters, we maximized the Area Under the Curve (AUC) for each fold. After deriving the AUC for every fold, we calculated the mean and standard deviation of these values over all folds of the k-fold cross-validation. The hyperparameters yielding the highest average validation AUC across all folds were deemed optimal.

Once the model was trained using optimized hyperparameters over the entire training cohort, we assessed its performance on a holdout test set using bootstrapping with replacement. We created 100 synthetic bootstrapped versions of the holdout test set. These bootstrapped test sets maintained the same length as the original test set. The model's performance was then calculated on all bootstraps. We reported the median and 95% confidence interval based on the bootstraps of several performance metrics like AUC, precision, sensitivity, specificity, and F1-score.

6## III. RESULTS

*A. Participants*

Table I shows the demographic and clinical variables distribution for the patients included in the analyses. Table II shows the distribution of the demographics stratified by stable and unstable conditions. The participants were divided into two cohorts: development and test set. The development set was used to train and tune our deep-learning methods, and the test set was used to evaluate the approaches. The majority of the participants in both cohorts were identified as White, but the proportion of White participants was higher in the development cohort (81.66%) than in the test cohort (65.38%). Hispanic patients were similarly represented in both groups, with 8.66% in the development cohort and 11.53% in the test cohort. The development cohort also had a higher percentage of female participants (33.33%) than the test cohort (16.66%). The average age of participants in the development cohort was 57.78 years, with a standard deviation (SD) of 16.94, while that of the participants in the test cohort was 53.88 years (SD 17.49), with a p p-value of 0.288. Among the participants' physical attributes in the two cohorts, both cohorts had similar average heights (p-value 0.648). Still, the average weight in the development cohort (89.19 kgs) was notably more than the test cohort (78.10 kgs), with a p-value of 0.054. The length of stay of the participants in the development cohort was 19 days (23 SD) and 30 days in the test cohort (48 SD), with a p-value of 0.053. Regarding the disease prevalence, several notable differences were observed: a marked increase in the occurrence of cancer in the test cohort, an almost double prevalence of renal diseases in the test cohort (26.92%) compared to the development cohort (13.33%), diabetes was more prevalent in the test cohort and liver-related diseases were nearly equally prevalent in both groups.

TABLE I
PATIENTS CHARACTERISTICS

| Variables | Development Cohort (N=60) | test Cohort (N=26) |
|---|---|---|
| Female sex. N (%) | 20 (33.33%) | 10 (16.66%) |
| Hispanic Ethnicity, N (%) | 5 (08.66%) | 3 (11.53%) |
| Age in years, mean (SD) | 57.78 (16.94) | 53.88 (17.49) |
| Height in cm, mean (SD) | 173.78 (9.26) | 172.80 (08.32) |
| Weight in kgs, mean (SD) | 89.19 (25.00) | 78.10 (16.33) |
| Length of stay in days, mean (SD) | 19.76 (23.56) | 30.23 (48.53) |
| Race: Number (Percentage of cohort) | | |
| White | 49 (81.66%) | 17 (65.38%) |
| African American | 8 (13.33%) | 4 (15.38%) |
| Other | 3 (00.05%) | 4 (15.38%) |
| Diseases: Number (Percentage of cohort) | | |
| Cancer | 1 (01.66%) | 4 (15.38%) |
| Cerebro-vascular | 8 (13.33%) | 4 (15.38%) |
| Dementia | 2 (03.33%) | 2 (07.69%) |
| Paraplegia Hemiplegia | 9 (15.00%) | 1 (03.84%) |
| Congestive Heart Failure | 7 (11.66%) | 2 (07.69%) |
| Chronic Obstructive Pulmonary Disease | 5 (08.66%) | 2 (07.69%) |
| Diabetes | 7 (11.66%) | 5 (19.23%) |
| Metastatic Carcinoma | 0 (00.00%) | 0 (00.00%) |
| Liver | 14 (23.33%) | 6 (23.07%) |
| Peptic Ulcer | 2 (03.33%) | 1 (03.84%) |
| Renal | 8 (13.33%) | 7 (26.92%) |

7TABLE II
DISTRIBUTION OF DEMOGRAPHICS VARIABLES STRATIFIED BY CLASS LABELS (STABLE, UNSTABLE)

| Variables | Stable Patients (N=82) | Unstable Patients (N=28) |
|---|---|---|
| Female sex, N (%) | 29 (35,36%) | 7 (25,00%) |
| Hispanic Ethnicity, N (%) | 8 (09.75%) | 1 (04,00%) |
| Age in years, mean (SD) | 56.86 (17.39) | 58.32 (16.11) |
| Height in cm, mean (SD) | 173.27 (9.04) | 174.69 (7.49) |
| Weight in kgs, mean (SD) | 85.82 (23.50) | 87.76 (25.54) |
| Length of stay in days, mean (SD) | 21.93 (32.78) | 25.92 (20.98) |
| Race | | |
| White, N (%) | 63 (76.82%) | 22 (78.57%) |
| African American, N (%) | 11 (13.41%) | 4 (14.28%) |
| Other, N (%) | 8 (09.75%) | 3 (10.71%) |

## B. *Experimental results*

We evaluated the performance of five deep learning models on different combinations of feature sets: accelerometer data only (Accel), accelerometer data with demographics (Accel + Demo), accelerometer data with clinical information (Accel + Clinical), and a combination of accelerometer data, demographics, and clinical information (Accel + Demo + Clinical). In our work, we refer as *demographics* the features of age, sex, race, height, weight, and length of stay, and as *clinical data*, the features of blood pressure, heart rate, oxygen saturation (SpO2), pain score, Braden score, and cognitive status. In addition, we used the SOFA score as a baseline to compare performances across the rule-based and deep learning-based methods. The results are summarized in Table III.

TABLE III
THE BEST RESULTS REPORTED AS AVERAGE AND 95% CONFIDENCE INTERVAL IN EACH SCENARIO

| | AUC (95% CI.) | Precision (95% CI.) | Sensitivity (95% CI.) | Specificity (95% CI.) | F1-score (95% CI.) |
|---|---|---|---|---|---|
| SOFA score | 0.53 (0.48-0.58) | 0.23 (0.19-0.28) | 0.30 (0.22-0.38) | 0.76 (0.69-0.82) | 0.66 (0.61-0.72) |
| Accel | 0.50 (0.50-0.50) | 0.61 (0.53-0.67) | 0.00 (0.00-0.00) | 1.00 (1.00-1.00) | 0.68 (0.62-0.73) |
| **Accel + Demo** | **0.69 (0.63-0.75)** | **0.75 (0.71-0.80)** | **0.67 (0.57-0.77)** | **0.63 (0.58-0.67)** | **0.67 (0.63-0.71)** |
| Accel + Clinical | 0.59 (0.53-0.65) | 0.61 (0.53-0.67) | 0.00 (0.00-0.00) | 1.00 (1.00-1.00) | 0.68 (0.62-0.73) |
| Accel + Demo + Clinical | 0.42 (0.35-0.48) | 0.69 (0.64-0.75) | 0.37 (0.27-0.46) | 0.74 (0.70-0.79) | 0.68 (0.64-0.72) |

*Abbreviations. Accel - accelerometer data, demo - demographics (age, sex, race, height, weight, and length of stay), clinical - the clinical set of features (blood pressure, heart rate, spo2, pain score, Braden score, and brain status)*

Our baseline SOFA score-based predictor demonstrated the lowest precision value of 0.23 (0.19-0.28), a sensitivity value of 0.30 (0.22-0.38), and a F1-score of 0.66 (0.61-0.72). It highlighted the limitations of traditional scoring systems, such as the SOFA score, in effectively assessing patients' acuity. In contrast, our deep learning models exhibited distinct performance trends. The model solely relying on accelerometer data yielded the lowest performance, with an AUC of 0.50 (0.50-0.50). This outcome was anticipated, considering the inherent challenge in assessing a patient's acuity based solely on movement patterns, although mobility is associated with recovery patterns. In contrast, the model incorporating accelerometer data and demographics outperformed others, achieving the highest AUC of 0.69 (0.63-0.75), along with precision, sensitivity, specificity, and F1-score values of 0.75 (0.71-0.80), 0.67 (0.57-0.77), 0.63 (0.58-0.67), and 0.67 (0.63-0.71) respectively.

We tuned the hyperparameter with Optuna, an optimization framework. Table IV outlines the best hyperparameters found by the search for each combination of feature sets. Appendix Table I provides a comprehensive overview of the hyperparameters and their corresponding values. The best-performing model for the scenario using only accelerometer data was MobileNet, with a batch size of 8, learning rate of $2.11 \times 10^{-4}$, weight decay of $1.51 \times 10^{-4}$, and an accelerometer downsampling factor of 1. The small batch size and relatively low learning rate allowed the model to fine-tune its weights, but the information coming from only accelerometer data was a limiting factor. In the Accel + Demo scenario, SENet showed the best performance with a batch size of 8, a learning rate of $2.34 \times 10^{-4}$, a weight decay of $1.41 \times 10^{-4}$, and a downsampling factor of 1. For the Accel + Clinical Data scenario, the Transformer model with a batch size of 32, a learning rate of $7.96 \times 10^{-2}$, a weight decay of $9.86 \times 10^{-5}$, and an

8accelerometer downsampling factor of 1 performed the best. The same Transformer model configuration also yielded the best results for the Accel + Demo + Clinical Data scenario but with a smaller batch size of 24 and a learning rate of $6.44 \times 10^{-3}$.

TABLE IV
BEST HYPERPARAMETERS FOR EACH SCENARIO

|  | Model | Batch size | Learning Rate | Weight Decay | Accelerometer Downsampling Factor |
|---|---|---|---|---|---|
| Accel | MobileNet | 8 | $2.11 \times 10^{-4}$ | $1.51 \times 10^{-4}$ | 1 |
| Accel + Demo | SENet | 8 | $2.34 \times 10^{-4}$ | $1.41 \times 10^{-4}$ | 1 |
| Accel + Clinical Data | Transformer | 32 | $7.96 \times 10^{-2}$ | $9.86 \times 10^{-5}$ | 1 |
| Accel + Demo + Clinical Data | Transformer | 24 | $6.44 \times 10^{-3}$ | $5.97 \times 10^{-4}$ | 1 |

*C. Discussion*

This study explored the potential of accelerometer and EHR data in directly determining patients' acuity levels, as an alternative to depending exclusively on rule-based scoring systems like SOFA. Our proposed methods achieved impressive performance when fusing the motion with patient contextual data, while accelerometer-only models and the fusion of the accelerometer data with EHR data had reduced performance.

The higher performance when combining accelerometer data with patients' demographics indicated that integrating contextual information about the patient with their mobility patterns significantly enhanced the model's ability to distinguish between stable and unstable conditions. Interestingly, the integration of clinical data did not result in a proportional enhancement of the model's performance, as demonstrated by the outcomes observed in the Accel + Clinical and Accel + Demo + Clinical scenarios. This discrepancy may be explained by the aggregation methodologies employed in this investigation, wherein temporal sequences were reduced to singular value features through processes such as averaging or majority voting. Such approaches might not be ideally suited for the task at hand. In the Accel + Demo scenario, SENet showed the best performance. SENet is a network designed to recalibrate feature maps adaptively and might have benefitted from the added context that demographics offered. The similar batch size and learning rate to the MobileNet model suggested that the added demographic features provide a richer representation that SENet could exploit.

Leveraging Optuna, we investigated configurations spanning various models, batch sizes, learning rates, weight decay, and accelerometer downsampling factors. The Accelerometer downsampling factor stands for evaluating the reduction in the sampling frequency of the accelerometer data. Given that the window size was set at 4 hours, this adjustment aimed to mitigate potential difficulties the models might encounter when processing these long sequences. This approach ensured a thorough examination of potential combinations, aiming to identify the most optimal setup for our specific dataset and problem context.

All models performed best with an accelerometer downsampling factor of 1, which indicated that the long sequence was not a problem for the employed architectures. Notably, a bigger batch size was necessary for the models when clinical data was included in the model. This could indicate that the added complexity and dimensionality introduced by the clinical data required more samples to be processed simultaneously for the model to effectively identify patterns, optimize the gradients, and achieve better convergence during training. Interestingly, Transformers demonstrated superiority in scenarios where clinical data was incorporated, even if not in a time-series format. Their inherent self-attention mechanism adeptly manages diverse data inputs, seamlessly integrating non-sequential clinical information with sequential accelerometer data.

This study was limited by the use of data from a single institution. Additionally, we summarized each clinical feature into a single value, which could minimize its contribution to the performance. Future work will incorporate EHR clinical data as a time series since the transformer model may potentially capture more of the intricate patterns and temporal relationships between the patient's physiological state and their movements, leading to more accurate and insightful predictions. Moving forward, our objective is to incorporate raw time-series clinical data directly into the model. Furthermore, we intend to leverage explainability algorithms such as SHAP analysis [26] and the LIME model [27] to delineate the contribution of individual features to the model's output, thereby facilitating a more granular understanding of the relationship between clinical variables and the resultant predictions.

## IV. CONCLUSION

Critical care environments necessitate the timely assessment of patient acuity to determine the severity of illness and prioritize care accordingly. Traditional methodologies have relied upon manual scoring systems, such as APACHE, SAPS, SOFA, and MEWS, which predominantly focus on physiological parameters. With the advent of technology, there has been a shift towards utilizing Electronic Health Records (EHR) integrated with machine learning techniques for such assessments. However, relying solely on physiological data captured within the EHR might overlook other crucial patient-related factors. This research aimed to assess the feasibility of combining accelerometer and EHR data for direct patient acuity assessment. The objective was to discern moments of stability and instability in a patient's condition. The study utilized five deep learning methods: VGG, ResNet, MobileNet, SqueezeNet, and a transformers-based network. Data was collected from June 2021 to February 2023 from 86 consenting patients with wrist-worn accelerometers. Relevant clinical data was extracted from the EHR. The results indicate that models relying solely on accelerometer data faced challenges, with an AUC of 0.50. However, when demographic information was integrated with accelerometer data, a notable performance improvement was attained, achieving an AUC of 0.69. It highlights the importance of a combined approach, wherein mobility patterns are complemented with contextual patient data. The introduction of additional clinical data did not yield significant improvements, suggesting that the manner in which the data is integrated–rather than the volume of data–is paramount to performance. This research emphasizes the significance of a comprehensive approach to patient acuity assessment in critical care settings. While initial results are promising, further research is needed to optimize the accuracy and efficiency of these assessments to ensure improvements in patient care and safety.

APPENDIX

TABLE I
OVERVIEW OF THE HYPERPARAMETERS AND THEIR RESPECTIVE VALUES EXPLORED IN THE HYPERPARAMETER OPTIMIZATION

| Hyperparameter | Values |
|---|---|
| Model | VGG, ResNet, MobileNet, SqueezeNet, and Transformers |
| Batch size | 8, 16, 24 and 32 |
| Learning Rate | ranging from $10^{-5}$ to $10^{-1}$ |
| Weight decay | ranging from $10^{-10}$ to $10^{-3}$ |
| Accelerometer downsampling factor | 1, 2 and 4 |